\documentclass{wscpaperproc}
\usepackage{latexsym}
\usepackage{graphicx}
\usepackage{mathptmx}

\usepackage{amsmath}
\usepackage{amsfonts}
\usepackage{amssymb}
\usepackage{amsbsy}
\usepackage{amsthm}
\usepackage{xcolor}

\usepackage{subfig}
\usepackage{floatrow} 
\newcommand{\R}{\mathbb{R}}
\newcommand{\E}{\mathbb{E}}

\newcommand{\X}{\mathbb{X}}
\newcommand{\W}{\mathbb{W}}
\newcommand{\EI}{\mathrm{EI}}
\newcommand{\KG}{\mathrm{KG}}

\newcommand{\savelength}[1]{} 

\DeclareMathOperator*{\argmax}{argmax}

\usepackage{color-edits} 
\addauthor{pf}{blue}
\addauthor{ra}{magenta}
\addauthor{mf}{red}

\usepackage[pdftex,colorlinks=true,urlcolor=blue,citecolor=black,anchorcolor=black,linkcolor=black]{hyperref}

\newtheoremstyle{wsc}
{3pt}
{3pt}
{}
{}
{\bf}
{}
{.5em}
{}

\theoremstyle{wsc}

\begin{document}

%
%

\pagestyle{fancyplain}

\thispagestyle{plain}
\firstPageHead{}

\chead{\fancyplain{}{\itshape Astudillo and Frazier}}

\rhead{}
\cfoot{}
\renewcommand{\headrulewidth}{0pt} 

\makeatletter
\let\@internalcite\cite
\def\cite{\def\@citeseppen{-1000}%
    \def\@cite##1##2{(##1\if@tempswa , ##2\fi)}%
    \def\citeauthoryear##1##2##3{##1 ##3}\@internalcite}
\def\citeNP{\def\@citeseppen{-1000}%
    \def\@cite##1##2{##1\if@tempswa , ##2\fi}%
    \def\citeauthoryear##1##2##3{##1 ##3}\@internalcite}
\def\citeN{\def\@citeseppen{-1000}%
    \def\@cite##1##2{##1\if@tempswa, ##2)\else{}\fi}%
    \def\citeauthoryear##1##2##3{##1 (##3)}\@citedata}
\def\citeA{\def\@citeseppen{-1000}%
    \def\@cite##1##2{(##1\if@tempswa , ##2\fi)}%
    \def\citeauthoryear##1##2##3{##1}\@internalcite}
\def\citeANP{\def\@citeseppen{-1000}%
    \def\@cite##1##2{##1\if@tempswa , ##2\fi}%
    \def\citeauthoryear##1##2##3{##1}\@internalcite}
\def\shortcite{\def\@citeseppen{-1000}%
    \def\@cite##1##2{(##1\if@tempswa , ##2\fi)}%
    \def\citeauthoryear##1##2##3{##2 ##3}\@internalcite}
\def\shortciteNP{\def\@citeseppen{-1000}%
    \def\@cite##1##2{##1\if@tempswa , ##2\fi}%
    \def\citeauthoryear##1##2##3{##2 ##3}\@internalcite}
\def\shortciteN{\def\@citeseppen{-1000}%
    \def\@cite##1##2{##1\if@tempswa, ##2\else{}\fi}%
    \def\citeauthoryear##1##2##3{##2 (##3)}\@citedata}
\def\shortciteA{\def\@citeseppen{-1000}%
    \def\@cite##1##2{(##1\if@tempswa , ##2\fi)}%
    \def\citeauthoryear##1##2##3{##2}\@internalcite}
\def\shortciteANP{\def\@citeseppen{-1000}%
    \def\@cite##1##2{##1\if@tempswa , ##2\fi}%
    \def\citeauthoryear##1##2##3{##2}\@internalcite}
\def\citeyear{\def\@citeseppen{-1000}%
    \def\@cite##1##2{(##1\if@tempswa , ##2\fi)}%
    \def\citeauthoryear##1##2##3{##3}\@citedata}
\def\citeyearNP{\def\@citeseppen{-1000}%
    \def\@cite##1##2{##1\if@tempswa , ##2\fi}%
    \def\citeauthoryear##1##2##3{##3}\@citedata}
%
%
%
\def\@citedata{%
    \@ifnextchar [{\@tempswatrue\@citedatax}%
                  {\@tempswafalse\@citedatax[]}%
}

\def\@citedatax[#1]#2{%
\if@filesw\immediate\write\@auxout{\string\citation{#2}}\fi%
  \def\@citea{}\@cite{\@for\@citeb:=#2\do%
    {\@citea\def\@citea{, }\@ifundefined
       {b@\@citeb}{{\bf ?}%
       \@warning{Citation `\@citeb' on page \thepage \space undefined}}%
{\csname b@\@citeb\endcsname}}}{#1}}%

%
\def\@citex[#1]#2{%
\if@filesw\immediate\write\@auxout{\string\citation{#2}}\fi%
  \def\@citea{}\@cite{\@for\@citeb:=#2\do%
    {\@citea\def\@citea{; }\@ifundefined
       {b@\@citeb}{{\bf ?}%
       \@warning{Citation `\@citeb' on page \thepage \space undefined}}%
{\csname b@\@citeb\endcsname}}}{#1}}%

%
\def\@biblabel#1{}
\makeatother



\newdimen\bibindent
\bibindent=0.0em
\def\thebibliography#1{\section*{\refname}\list
   {}{\settowidth\labelwidth{[#1]}
   \leftmargin\parindent
   \itemindent -\parindent
   \listparindent \itemindent
   \itemsep 0pt
   \parsep 0pt}
   \def\newblock{}
   \sloppy
   \sfcode`\.=1000\relax}


\setlength{\baselineskip}{12.7pt}

\title{THINKING INSIDE THE BOX: A TUTORIAL ON GREY-BOX BAYESIAN OPTIMIZATION}

\author{Raul Astudillo \& Peter I. Frazier\\[12pt]
School of Operations Research and Information Engineering\\
Cornell University\\
Ithaca, NY 14853, USA\\
}

\maketitle
\section*{ABSTRACT}
Bayesian optimization (BO) is a framework for global optimization of expensive-to-evaluate objective functions. Classical BO methods assume that the objective function is a black box. However, internal information about objective function computation is often available. For example, when optimizing a manufacturing line's throughput with simulation, we observe the number of parts waiting at each workstation, in addition to the overall throughput. Recent BO methods leverage such internal information to dramatically improve performance. We call these ``grey-box" BO methods because they treat objective computation as partially observable and even modifiable, blending the black-box approach with so-called ``white-box" first-principles knowledge of objective function computation. This  tutorial  describes these methods, focusing on BO of composite objective functions, where one can observe and selectively evaluate individual constituents that feed into the overall objective; and multi-fidelity BO, where one can evaluate cheaper approximations of the objective function by varying parameters of the evaluation oracle.

\section{INTRODUCTION}
\label{sec:intro}
Bayesian optimization (BO) is a framework for global optimization of objective functions that are expensive or time-consuming to evaluate.  The standard BO problem is of the form
$\max_{x\in\X}f(x)$,
where $f$ is an expensive-to-evaluate continuous function and $\X\subset \R^d$ is a simple compact set such as a hyperrectangle or a polytope.
Classical BO methods make no other explicit assumptions on the objective function.

The main characteristic of the BO paradigm is that $f$ is modeled as a realization from a Bayesian prior distribution over functions, with Gaussian processes being the most widely used family of distributions. 
Within an iterative algorithm,  this prior distribution along with the evaluations of $f$ performed so far give rise to a posterior distribution which is used via an acquisition function that quantifies the value of information from an objective function evaluation to 
select the next point at which to evaluate $f$. 

These methods are appealing because they can be easily applied without detailed knowledge of the objective function or derivative evaluations, in contrast with classical nonlinear  optimization methods, but nonetheless perform reasonably well across a wide variety of problems \shortcite{calandra2016bayesian,turner2021bayesian}. At the same time, they are flexible and permit the introduction of prior information from domain experts in the form of an informative prior distribution, in contrast with non-Bayesian derivative-free methods  \shortcite{conn2009introduction}.

BO originated with the seminal works of
\shortciteN{kushner1964}, \shortciteN{movckus1975bayesian}, and
\shortciteN{zhilinskas1975single}, focused on engineering design,
but is best known for its recent success in hyperparameter tuning of machine learning algorithms \shortcite{snoek2012practical,swersky2013multi,wu2019practical}. Beyond engineering design and hyperparameter tuning, BO has also been successful in many  other application areas, including 
operations-focused optimization via simulation applications
\shortcite{pearce2017bayesian},
drug discovery \shortcite{griffiths2020constrained}, 
and robotics \shortcite{calandra2016bayesian}.

While BO has been broadly successful,
the expense of evaluations nevertheless remains prohibitive
in a number of problem domains.
For example, consider optimizing population-level interventions such as social distancing and masking to prevent the spread of a disease based on predictions from an agent-based simulator that models the detailed movements of millions of people.
Suppose each evaluation takes several hours on a high-performance computing cluster, the search domain is 10-dimensional, and there are many local maxima. In such problems, it is plausible that a black-box method would need thousands of evaluations or more before it finds a solution close to the global optimum, requiring months of computation.
At the same time, suppose this 
agent-based simulation can be run with a smaller population size 
to obtain an approximation to the objective in dramatically less time. \savelength{(The ``mesh size'' describes the level of discretization used to solve the PDE. A larger mesh size corresponds to more discrete points used in the discretization, a more accurate solution, and more computation.)}
Then, it becomes appealing to use a method that uses such less accurate but faster approximations to understand how the objective behaves at a high level (see, e.g., Figure~\ref{fig:mf_example}), and only afterward focuses its attention on high-value regions.

This approach, in which lower fidelity faster evaluations are used in concert with higher fidelity slow evaluations, is called {\it multi-fidelity  optimization} (or \textit{multi-fidelity  BO} when specialized to the BO setting) \shortcite{huang2006sequential,forrester2007multi}.
Such methods generate value by sacrificing the generality of black box optimization, leveraging knowledge and access to the internals of the objective function evaluation to \savelength{significantly} accelerate search. While such methods require more specialization, they can be much faster.

\begin{figure}[h]
\floatbox[{\capbeside\thisfloatsetup{capbesideposition={right,top},capbesidewidth=8cm}}]{figure}[\FBwidth]
{\caption{
Performance of  grey-box BO  ($\EI_{\mathrm{gb}}$) compared to standard black-box BO ($\EI_{\mathrm{bb}}$) and random search ($\mathrm{Random}$)  on the  calibration test problem described in \S5.3 of Astudillo and Frazier (2019). Grey-box BO leverages the composite structure of the objective function and, by doing so, it dramatically improves performance. Grey-box BO achieves the same regret as standard BO after 40 evaluations using only 5 evaluations. 
\label{fig:gbei_vs_bbei}}}
{\includegraphics[width=0.4\textwidth]{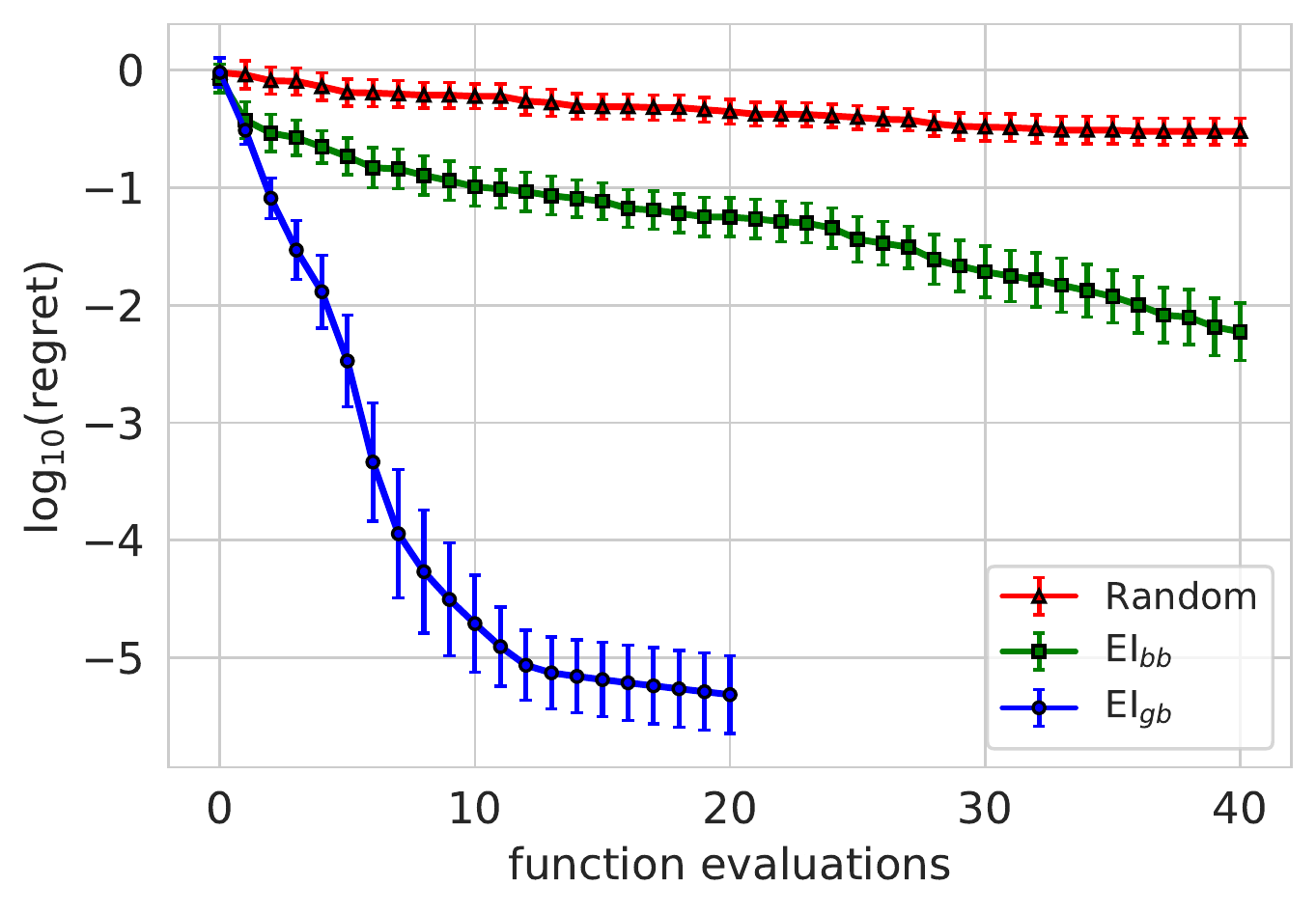}}
\end{figure}

As we articulate here, multi-fidelity BO is just one example within a broader class of methods that leverage knowledge and access to the internals of objective function evaluation to improve efficiency. We refer collectively to such methods as {\it grey-box Bayesian optimization methods}.
Specifically, we refer to any method as a grey-box BO method if it leverages access to the internal computational structure of objective function or constraint evaluation. 
This can deliver dramatic performance gains, sometimes 
improving accuracy multiple orders of magnitude
 at a given level of computational effort (see, e.g., Figure~\ref{fig:gbei_vs_bbei}).
 
Existing work on grey-box BO can be broadly divided into three classes: BO of composite objective functions; multi-fidelity BO; and BO with objective function constituent evaluations.

\begin{itemize}
\item \textbf{BO of composite objective functions} 
uses observations of 
internal constituents  that  feed  into  the  overall  objective value calculation to improve the predictive model of the objective function.
This arises, for example, when calibrating a simulator's parameters to data and in inverse reinforcement learning.
The objective function at a given vector of parameters $x$ is the sum of squared errors between the simulator's prediction $h_j(x)$ for an experimental condition and an observation $y_j^{\mathrm{obs}}$ of that condition. Rather than simply modeling the overall objective as one monolithic black box, one can model each function $h_j$  as a black box and understand that the objective is the composition of these functions with the function $g(y) = \sum_j (y_j-y_j^{\mathrm{obs}})^2$. Using observations of $h_j(x)$ provides substantially more information that can be used to select good $x$ at which to evaluate
\shortcite{uhrenholt2019efficient,astudillo2019bayesian}.
This also arises, for example, \savelength{ inmulti-disciplinary optimization \shortcite{cramer1994problem};} in aerospace engineering, where multiple physics-based simulators pass information back and forth, creating an objective function that is a composition of a collection of black-box functions; 
\savelength{tuning hyperparameters of a machine learning method to minimize cross-validation error \shortcite{swersky2013multi} (the objective is the average of errors associated with each individual fold);}
and when minimizing a cost function that aggregates a simulator's predictions across a variety of environmental conditions.

\item \textbf{Multi-fidelity BO} modifies the process of evaluating the objective function so that the output is not the objective value itself but is instead a faster-to-compute but less accurate approximation.
Modifications to the objective function evaluation process include using smaller mesh sizes when solving partial differential equations (PDEs), reducing run-lengths when the objective function is the output of a steady-state simulation, and reducing the number of training iterations when the objective is the test error for a deep neural network (DNN).
This is the most widely studied class of grey-box methods to date.

\item \textbf{BO with objective constituent evaluations}
leverages the ability to evaluate just some of the constituents that make up the objective function to save time while also enabling learning, either delaying the evaluation of other constituents until some future point in time or not evaluating them at all. This is possible in both BO of composite objective functions and multi-fidelity BO.
For example, when minimizing average cost over scenarios, one can evaluate cost on a subset of the possible scenarios; or when minimizing the test error for a DNN, an evaluation can be paused at a small number of training iterations (and continued later if desired).
\end{itemize}

The term ``grey-box'' originates from physics-based modeling, where black-box models are purely empirical models that include no first-principles theoretical knowledge from physics, white-box models exclusively rely on such detailed physical knowledge, and grey-box models blend the two approaches \shortcite{tullekengrey1993,bohlin2006practical}. In grey-box BO, we adopt this same terminology, except that we seek to model the objective function rather than the real world. We use a blend of empirical data-driven methods (black-box methods) and detailed first-principles knowledge of how the objective is computed (white-box methods). Distinct from grey-box optimization, there is work applying black-box surrogate-based optimization methods (which assume nothing about the structure of the objective function and constraints) to optimize such grey-box physics-based models \shortcite{beykal2018global}.

To help clarify the use of the term grey-box optimization, we also mention here other methods that go beyond standard BO but that we do not consider to be grey-box BO: 
\begin{itemize}
\item BO with shape constraints \shortcite{jauch2016bayesian}. This uses knowledge of the objective function but does not leverage access to its internal computation.
\item High-dimensional BO assuming additive structure 
\shortcite{gardner2017discovering}
or a linear embedding in a low-dimensional space \shortcite{letham2020re}, when such structure is used or assumed (as it often is) without access to the internals of objective function calculation.
\item BO with gradient information \shortcite{wu2017bayesian}. Gradient information is commonly included in objective function oracles used in classical (non-Bayesian) optimization, and is viewed as an externally-provided output. We take that view here.
\end{itemize}

\paragraph{Organization of the rest of this tutorial} The rest of this tutorial first provides a brief introduction and basic concepts in standard BO in \S\ref{sec:standard-BO}.  For a more detailed tutorial on standard BO, we refer the reader to \shortciteN{frazier2018tutorial}. This tutorial on grey-box BO will be most enjoyable to those who have already read such a tutorial focused on standard BO.
Then, \S\ref{sec:composite-BO}-\ref{sec:constituent-BO}
describe different types of grey-box BO:
\S\ref{sec:composite-BO} describes BO of composite objective functions;
\S\ref{sec:multifidelity-BO} describes multi-fidelity BO.
and \S\ref{sec:constituent-BO} describes BO with objective constituent evaluations.
\S\ref{sec:conclusion} concludes while offering directions for future research.

\savelength{
The goal of this tutorial is to elucidate and explain these three classes of problems, and to position researchers interested in this area to rapidly advance the state of the art.
Grey-box BO is an exciting area where a great deal remains to be discovered, with numerous opportunities to leverage the internal computational structure of objective / constraint evaluation to dramatically improve efficiency.
}

\section{STANDARD BAYESIAN OPTIMIZATION} 
\label{sec:standard-BO}
A BO method consists of two main components: a predictive model, given by a Bayesian prior distribution over $f$ that serves as a surrogate equipped with uncertainty estimates; and an acquisition function, which depends on the implied posterior distribution over $f$ given the set of available evaluations so far, and whose value at an arbitrary point $x\in\X$ quantifies the \textit{benefit} of evaluating at this point. In this section, we discuss these two components in detail, focusing on Gaussian processes (GPs), which is the class of probability distributions most widely used in BO, and reviewing several popular acquisition functions. 
\subsection{Predictive Model}
As mentioned above, the first component of a BO method is a predictive model, given by a Bayesian prior probability distribution over $f$. Examples of probability distributions  used in the BO literature include random forests \shortcite{hutter2011sequential}, Bayesian neural networks \shortcite{snoek2015scalable} and GPs. Here, we focus on the latter class of probability distributions, which is arguably the most widely used in practice due to its computational tractability and well-calibrated uncertainty estimates. For a detailed discussion on GPs, we refer the reader to \shortciteN{rasmussen2006gaussian}.

A GP prior distribution over $f$ is fully determined by a prior mean function, $\mu_0 : \X \rightarrow, \R$ and a prior covariance function, $K_0 : \X \times \X \rightarrow \R$. It has the property that, for any finite collection of points  $x_1,\ldots, x_n \in \X$, the prior distribution of $(f(x_1),\ldots, f(x_n))^{\top}$ is multivariate normal with mean vector $(\mu_0(x_1),\ldots, \mu_0(x_n))^{\top}$ and covariance matrix $(K_0(x_i, x_j))_{i,j=1}^n$. Moreover, given a data set of $n$ (potentially noisy) evaluations, $\mathcal{D}_n = \{(x_i, y_i)\}_{i=1}^n$, where $y_i = f(x_i) + \epsilon_i$ and $\epsilon_1,\ldots, \epsilon_n$ are i.i.d. from a normal distribution with mean zero and variance $\sigma^2$, the posterior distribution on $f$ is again a GP with mean and covariance functions 
\begin{align*}
  \mu_n(x) &= \mu_0(x) + K_0(x, x_{1:n})\left[K_0(x_{1:n}, x_{1:n}) + \sigma^2 I_n\right]^{-1} (y_{1:n} - \mu_0(x_{1:n})),\\
    K_n(x,x') &= K_0( x, x') - K_0(x, x_{1:n})\left[K_0(x_{1:n}, x_{1:n}) + \sigma^2 I_n\right]^{-1}K_0( x_{1:n}, x'),
\end{align*}
respectively, where $I_n$ is the $n$-dimensional identity matrix and, making a slight abuse of notation, we define $y_{1:n} = (y_1,\ldots,y_n)^\top$, $\mu_0(x_{1:n}) = (\mu_{0}(x_1),\ldots, \mu_{0}(x_n))^\top$, $K_0(x_{1:n}, x_{1:n})=(K_0(x_i, x_j))_{i,j=1}^n$, $K_0(x, x_{1:n}) = (K_0(x,x_1), \ldots, K_0(x,x_n))$, and  $K_0(x_{1:n}, x') = (K_0(x_1,x'), \ldots, K_0(x_n, x'))^\top$.

Following the Bayesian statistics terminology, $\mu_n$ and $K_n$ are called the posterior mean and covariance functions, respectively.  The function $\mu_n$ can be interpreted as a surrogate for $f$, whereas $K_n$ equips this surrogate with uncertainty estimates. In particular, when evaluations of $f$ are noiseless; i.e., when $\sigma^2 = 0$, the posterior mean function interpolates the evaluations of $f$ collected so far and their corresponding uncertainty estimates are exactly 0; i.e., $\mu_n(x_i)=f(x_i)$ and $K_n(x_i, x_i)=0$ for $i=1,\ldots, n$.

\subsection{Acquisition Function}
The second component of a BO method is an acquisition function, $\alpha_n: \X \rightarrow \R$, where the sub-index $n$ indicates the dependence  on the posterior distribution on $f$ at time $n$. The value of the acquisition function at a particular point $x\in\X$ can be interpreted as a measure  of the \textit{benefit} of evaluating at this point, and thus one wishes to evaluate a point with the highest acquisition value possible. Formally, a BO method using an acquisition function $\alpha_n$ chooses the next point to evaluate, $x_{n+1}$, as a maximizer of $\alpha_n$; i.e.,
    $x_{n+1}\in \argmax_{x\in \X}\alpha_n(x)$.
Importantly, unlike $f$, $\alpha_n$ is not expensive-to-evaluate and its gradients are typically available. This makes optimizing the acquisition function substantially easier than optimizing $f$.

Acquisition functions navigate the trade-off between evaluating points whose objective value is still very uncertain and those whose objective value is expected to be high, commonly known as the exploration-exploitation trade-off. Popular acquisition functions include \savelength{probability of improvement (PI) \shortcite{kushner1964},} expected improvement (EI) \shortcite{zhilinskas1975single,jones1998efficient}, knowledge gradient (KG) \shortcite{frazier2008knowledge,frazier2009knowledge,scott2011correlated}, Gaussian process upper confidence bound (GP-UCB) \shortcite{srinivas2009gaussian}, entropy search (ES) \shortcite{hennig2012entropy}, predictive entropy search (PES) \shortcite{hernandez2014predictive}, and max-value entropy search \shortcite{wang2017max} (MVES).  Below, we discuss in detail  EI and KG due to their simplicity and also because they have been the most widely generalized to grey-box settings.

\subsubsection{Expected Improvement}
The most widely used acquisition function in standard BO is the expected improvement (EI), which is defined by
\begin{equation*}
    \EI_n(x) = \E_n[\{f(x) - f_n^*\}^+],
\end{equation*}
where the sub-index $n$ indicates that the expectation is taken under the posterior distribution at time $n$, and $f_n^* = \max_{i=1,\ldots, n}f(x_i)$ is the best observed objective value so far. Interpreting $f^*_n$ as the reward that would be received if we reported a solution to our optimization problem after $n$ evaluations, $f_{n+1}^* - f_n^* = \{f(x) - f_n^*\}^+$ is the improvement in this reward due to an additional sample at $x$.

The EI acquisition function was first proposed by \shortciteN{movckus1975bayesian} and popularized by \shortciteN{jones1998efficient}. It is known to perform well in practice, especially when evaluations are noiseless. At the same time, it is also known to be outperformed by other more sophisticated acquisition functions when evaluations are noisy \shortcite{frazier2018tutorial} or the objective function is highly multi-modal \shortcite{jiang2020binoculars}.

In addition to its good empirical performance, another property contributing to EI's popularity is that it admits an analytic expression when $f$ is modeled using a GP and evaluations are noiseless. This analytic expression, obtained by noting that $\{f(x) - f_n^*\}^+$ is a truncated normal random variable, is given by
\begin{equation*}
    \EI_n(x) = \Delta_n(x)\Phi\left(\frac{\Delta_n(x)}{\sigma_n(x)}\right) + \sigma_n(x)\varphi\left(\frac{\Delta_n(x)}{\sigma_n(x)}\right),
\end{equation*}
where $\Delta_n(x) = \mu_n(x) - f_n^*$, $\sigma_n(x) = \sqrt{K_{n}(x,x)}$, and $\varphi$ and $\Phi$ are the standard normal probability density function (PDF) and cumulative distribution function (CDF), respectively. Importantly, if $\mu_n$ and $K_n$ are differentiable, so is $\EI_n$, and thus (deterministic) gradient-based optimization methods can be used to maximize $\EI_n$. A common choice in practice is to use L-BFGS-B \shortcite{byrd1995limited} with multiple restarts.

\subsubsection{Knowledge Gradient}
The knowledge gradient (KG) acquisition function was proposed by \shortciteN{frazier2008knowledge} for Bayesian ranking and selection among a finite number of alternatives and later adapted to the BO setting by \shortciteN{scott2011correlated}. Since then, it has been generalized to handle parallel evaluations \shortcite{wu2016parallel}, derivative information \shortcite{wu2017bayesian}, multiple information sources \shortcite{poloczek2017multi}, and also adapted to multiple grey-box settings that we describe in detail in 
\S\ref{sec:composite-BO}, \S\ref{sec:multifidelity-BO}, and \S\ref{sec:constituent-BO}. 

KG modifies the reward in EI's definition.
EI assumes that the reward for reporting a solution to our optimization problem at time $n$ is the maximum objective value across evaluated points, $f^*_n$. 
KG instead assumes this reward is the maximum (expected) objective value across the whole feasible space, $\mu_n^* = \max_{x\in\X}\E_n[f(x)]=    \max_{x\in\X}\mu_n(x)$. The reward $f^*_{n}$ is only improved by full evaluations of the objective function, but $\mu^*_n$ can be improved by any information about the objective function. This allows KG to generalize easily to grey-box settings. Analogously to EI, KG is defined as the expected change in reward from one additional evaluation; i.e., 
\begin{equation*}
    \KG_n(x) = \E_n[\mu_{n+1}^* - \mu_n^* \mid x_{n+1}=x].
\end{equation*}

While $\mu_n^*$ is deterministic given the information available up to time $n$ (and thus can be pulled out of the expectation above), $\mu_{n+1}^*$ is random due to its dependence on the yet unobserved value of $y_{n+1}$. 
Moreover, when $\X$ is continuous, $\KG_n$ does not admit a simple analytic expression, and thus maximizing it requires solving a nested stochastic optimization problem. This makes KG significantly harder to maximize than acquisition functions with simple analytic expressions like EI. Very often, however, the extra computation required to maximize KG is justified by its superior performance. This is particularly true in grey-box settings where KG often admits natural generalizations, whereas other acquisition functions are adapted to these settings via less principled heuristics or approximations which often lead to a worse performance.

Several approaches have been  proposed in the literature to  maximize KG. \shortciteN{scott2011correlated} proposes to approximate $\KG_n$ by replacing $\mu_m^*$  by $\widetilde{\mu}_m^* = \max_{i=1,\ldots,n+1}\mu_m(x_i)$, for $m=n, n+1$. The key advantage of this approximation is that it admits an analytic expression which, like EI, can be maximized using deterministic gradient-based optimization methods. However, it becomes computationally burdensome for problems of moderate dimension. \shortciteN{wu2017bayesian} proposes an \textit{exact} approach to maximize KG by computing stochastic gradients of $\KG_n$ which are then used in  a multi-start stochastic gradient ascent (SGA) routine. This approach scales better to problems of moderate dimension. However, using SGA requires choosing its learning rate, which can be non-trivial.  \shortciteN{balandat2020botorch} proposes a \textit{one-shot optimization}  approach that effectively replaces the original problem of maximizing $\KG_n$ with a sample average approximation (SAA). This approximate problem is not only deterministic but it can also be cast as a non-nested optimization problem over a higher dimensional space, thus allowing again the use of deterministic gradient-based optimization methods. However, the dimension of this approximate problem grows linearly with the number of samples used, restricting the number that can be used in practice.

The above three approaches to maximize KG (implicitly or explicitly) rely on the so-called reparametrization trick for acquisition functions \shortcite{wilson2018maximizing}, which consists on rewriting an acquisition function as an expectation of a deterministic transformation (depending on the posterior mean and covariance functions) of a standard normal random variable. Such an approach has also been key to extending other acquisition functions such as EI or GP-UCB to settings where they no longer have an analytic expression such as batch evaluations \shortcite{wilson2018maximizing,wang2016parallel} and composite objective functions \shortcite{astudillo2019bayesian}. The reparameterized expression of KG is given by
\begin{equation*}
    \KG_n(x) = \E_n\left[ \max_{x'\in\X}\mu_n(x') + \widetilde{\sigma}_n(x'; x_{n+1})Z \mid x_{n+1}=x\right] - \mu_n^*,
\end{equation*}
where $\widetilde{\sigma}(x'; x_{n+1}) = K_n(x',x_{n+1})/\sqrt{K_n(x_{n+1}, x_{n+1}) + \sigma^2}$, and the (conditional) distribution of $Z$ is standard normal. We refer the reader to \shortciteN{frazier2009knowledge} and \shortciteN{wu2016parallel} for a derivation.

\section{BAYESIAN OPTIMIZATION OF COMPOSITE OBJECTIVE FUNCTIONS}
\label{sec:composite-BO}
The first grey-box BO setting we consider is the \textit{composite objective functions} setting, where the objective function is a known transformation of a vector-valued black-box function. Formally, we assume that the objective function, $f:\X\rightarrow\R$, is known to be of the form $f(x) = g(h(x))$, where $h:\X\rightarrow\R^k$ is a black-box expensive-to-evaluate vector-valued function, and $g:\R^k\rightarrow\R$ is a cheap-to-evaluate scalar function, typically known in closed form. This occurs, for example, in simulation calibration and inverse reinforcement learning, where $h(x)$ is a vector containing predictions for reality and the goal is to find the design variables $x$ so that these predictions most closely matches a vector data observed in the real world, $y^{\mathrm{obs}}\in\R^k$. In this case, a common choice is to minimize $f(x) = g(h(x))$, where $g(y)=\|y - y^{\mathrm{obs}}\|_2^2$.

\subsection{Predictive Model}
\begin{figure*}
\subfloat[\raggedright Standard black-box BO does not model $h$.]{%
\includegraphics[width=0.46\linewidth]{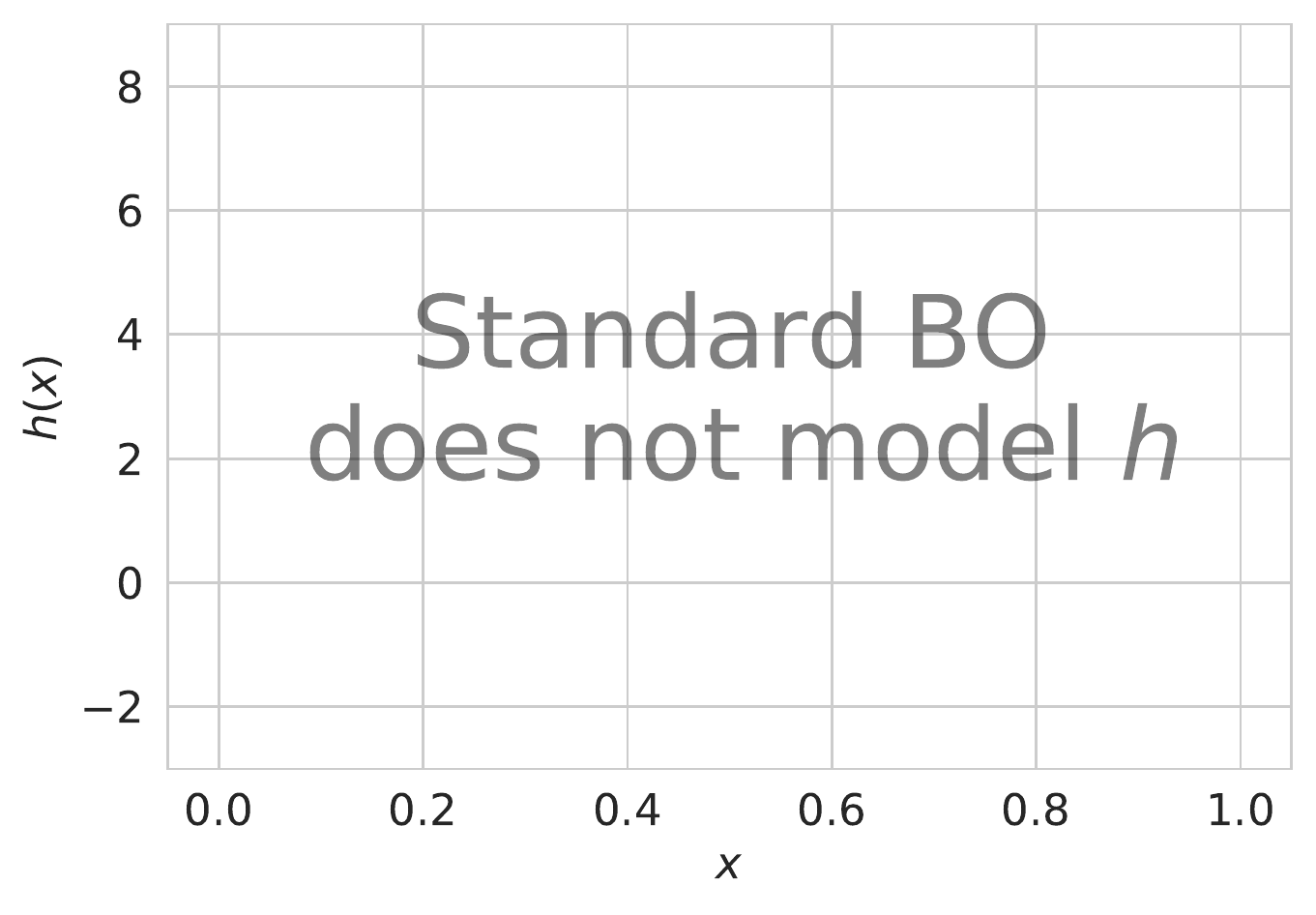}
}
\hfill
\subfloat[\raggedright GP posterior on $h$ used by grey-box BO. The optimum occurs when $h(x)=0$.]{%
\includegraphics[width=0.46\linewidth]{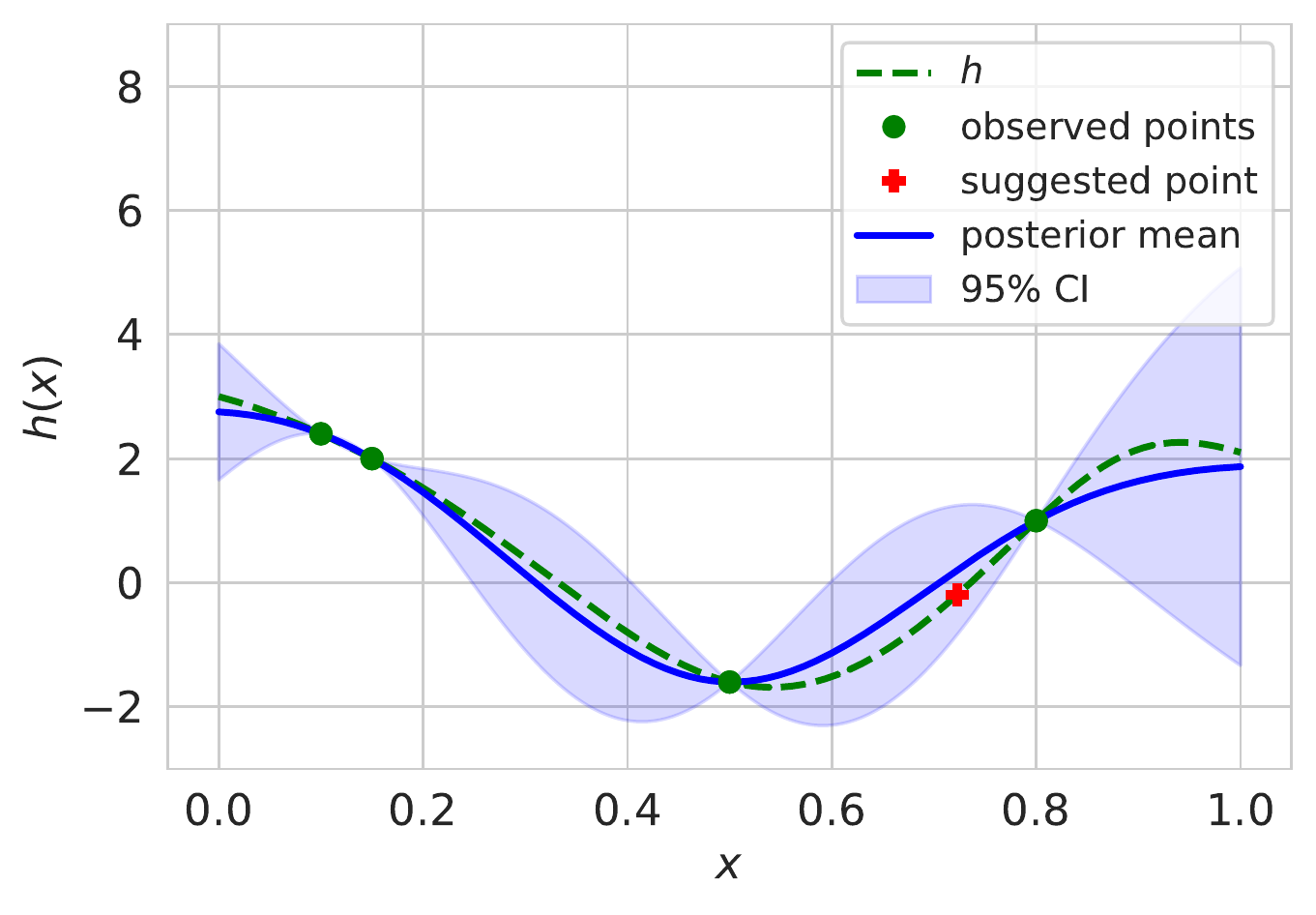}
}
\\
\subfloat[\raggedright GP posterior on $f$ used by standard black-box BO.]{%
\includegraphics[width=0.46\linewidth]{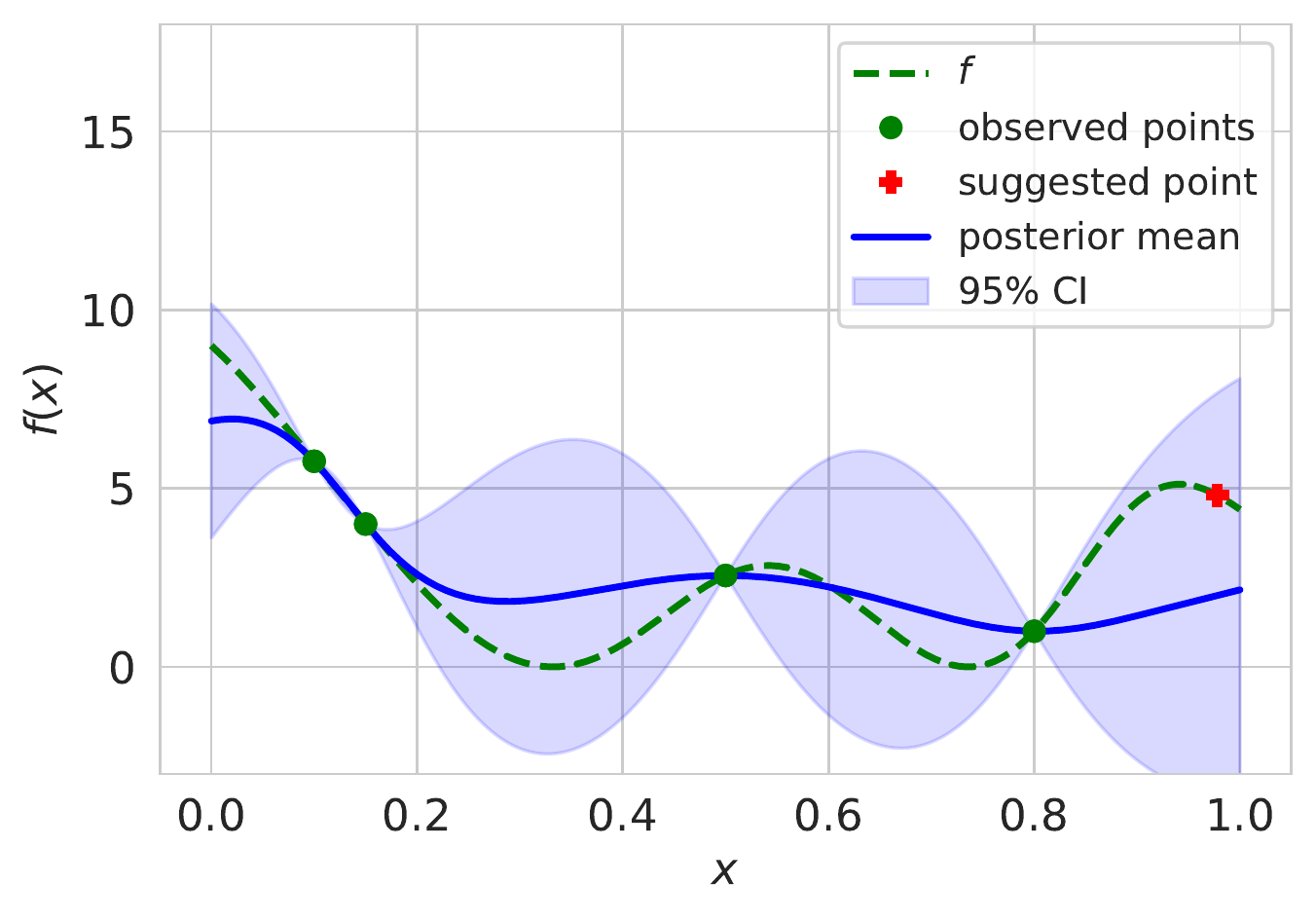}
}
\hfill
\subfloat[\raggedright Implied non-Gaussian posterior on $f$ used by grey-box BO. This posterior puts all mass on positive values.]{%
\includegraphics[width=0.46\linewidth]{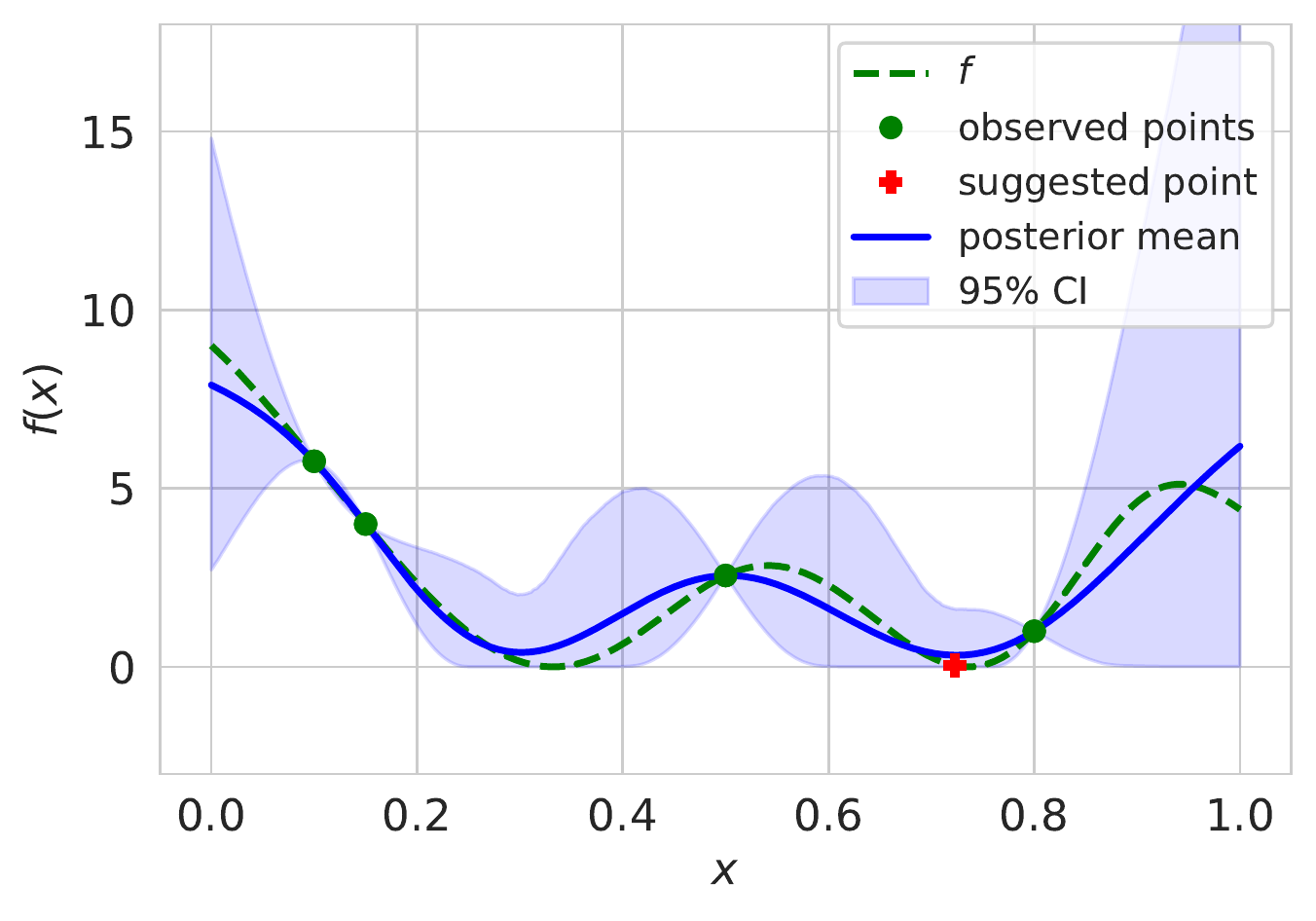}
}
\\
\subfloat[\raggedright EI computed with respect to the GP posterior on $f$ used by standard black-box BO.]{%
\includegraphics[width=0.46\linewidth]{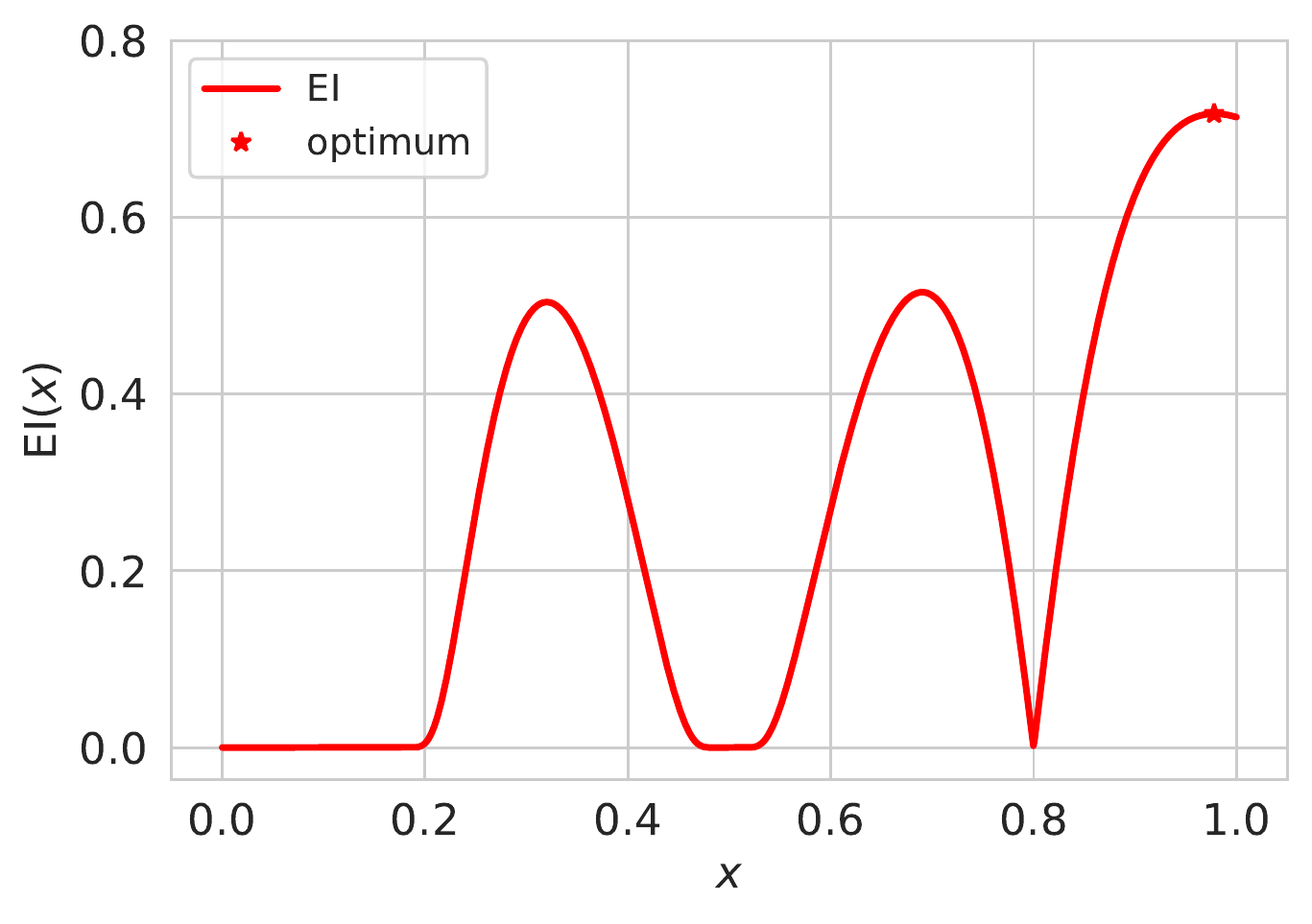}
}
\hfill
\subfloat[\raggedright EI computed with respect to the non-Gaussian posterior on $f$ used by grey-box BO.]{%
\includegraphics[width=0.46\linewidth]{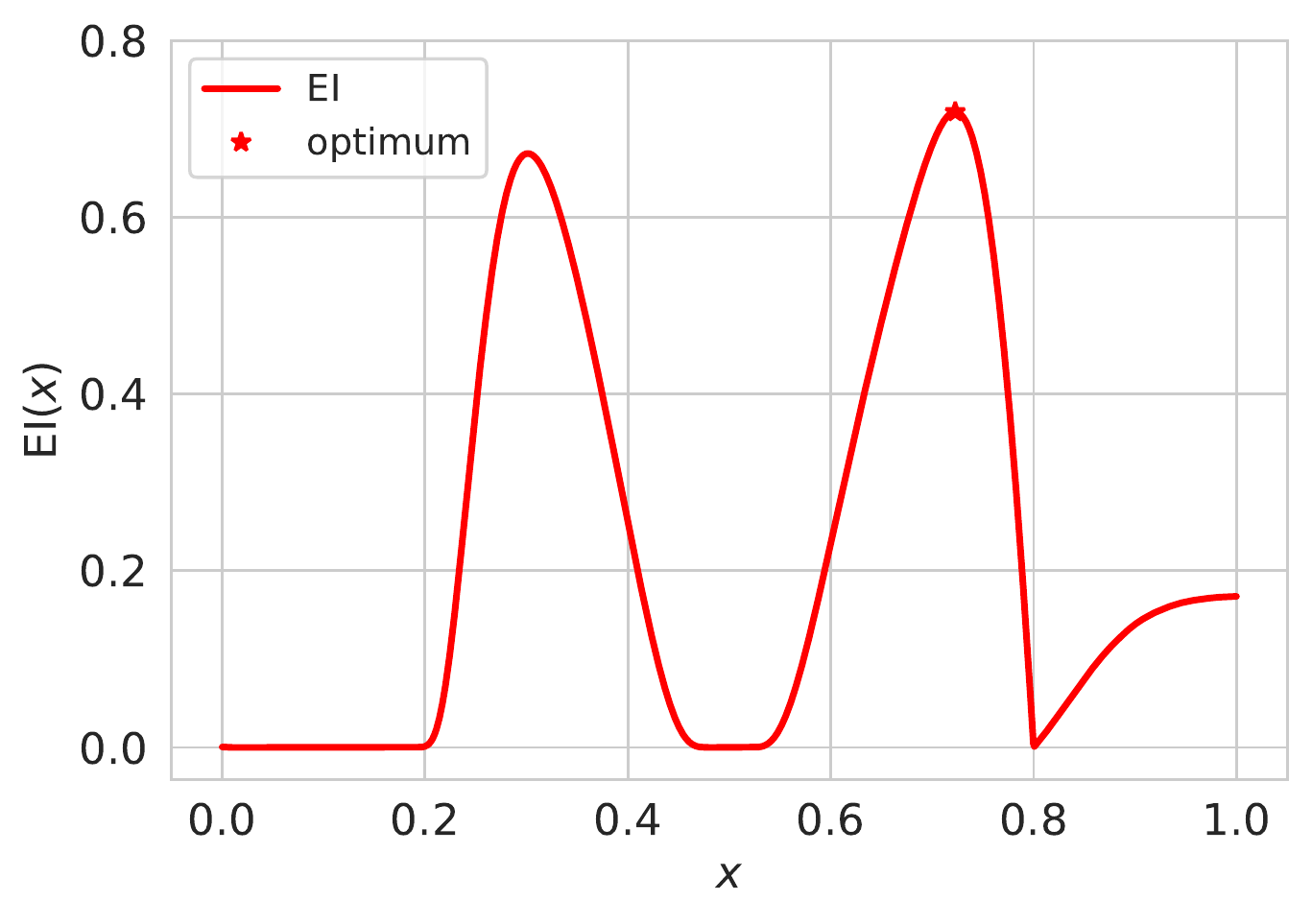}
}
\caption{Illustrative example of BO of a composite objective function in a (minimization) problem where $h$ is scalar-valued and $g(h(x))=h(x)^2$. 
Observations of $h(x)$ provide a substantially more accurate view of where global optima of $f$ reside as compared with observations of $f(x)$ alone.  This allows  grey-box BO  (right) to evaluate at points much closer to these global optima compared to standard black-box BO (left).
\label{fig:bocf_vs_bo}}
\end{figure*}

Notably, although evaluations of $h$ are available when computing the objective function, the standard BO approach does not use this information (directly). Intuitively, using this information can be beneficial, especially when $h$ carries information relevant for optimization that is not available from $f$ alone. For example, suppose we wish to minimize $f(x)=h(x)^2$, where $x$ and $h(x)$ are both scalars. If $h(x_1)< 0 < h(x_2)$ for some $x_1 < x_2$ and $h$ is continuous, then we know there exists $x^*\in(x_1,x_2)$ such that $h(x^*)=0$, making $x^*$  a global minimizer of $f$. This valuable information, however, is ignored by standard BO methods.  Figure \ref{fig:bocf_vs_bo} shows that, in this example, a grey-box BO method that explicitly models $h$ can indeed make a much better sampling decision than a standard black-box BO method that ignores it.

\shortciteN{astudillo2019bayesian} addresses this shortcoming by modeling $h$ using a multi-output GP, instead of  $f$ using a (single-output) GP as in the standard BO approach. Formally, this approach places a multi-output GP prior distribution on $h$ \shortcite{alvarez2012kernels}, which is again characterized by a prior mean function, $\mu_0:\X \rightarrow \R^k$, and a prior covariance function, $K_0:\X\times\X\rightarrow\R^{k\times k}$. Analogous to the single-output case, the posterior distribution on $h$ given $n$ of its evaluations is again a multi-output GP with posterior mean and covariance functions, $\mu_n$ and $K_n$, respectively, which can be computed in closed form. This posterior distribution on $h$ in turn implies a posterior distribution on $f$, which is in general non-Gaussian.

\subsection{Acquisition Functions}
Having specified a Bayesian prior probability distribution over $f$, it remains to specify an acquisition function.  When $g$ is non-linear, however, the posterior distribution over $f$ is no longer a GP. Therefore, classical acquisition functions such as PI, EI or GP-UCB no longer have a closed form, thus making them more challenging to compute and maximize.

\shortciteN{uhrenholt2019efficient} considers minimization of $g(h(x))=\|h(x) - y^{\mathrm{obs}}\|_2^2$, under which the implied posterior distribution on $f(x)$ is a generalized chi-squared distribution. The EI acquisition function under this distribution does not have a closed form expression. However, it is argued that, when the outputs of $h$ are modeled using independent GPs, this distribution can be well approximated by a scaled non-central chi-squared distribution with the same degrees of freedom ($k$) and non-centrality parameter, where the multiplying factor is chosen so that its expected value matches the one from the true distribution. Under this approximated distribution, EI has a closed-form analytical expression in terms of non-central chi-squared CDFs and can be efficiently optimized using deterministic gradient-based optimization methods. 

While the approach proposed by \shortciteN{uhrenholt2019efficient} is appealing due to the closed form analytical expression it provides, the performed approximation has unclear effects. Furthermore, it does not naturally extend to other functions $g$. \shortciteN{astudillo2019bayesian} addresses the more general case by noting that, for arbitrary $g$, the reparametrization trick can be used to rewrite $\EI_n$ as
\begin{equation}
\label{eq:ei_rep_trick}
    \EI_n(x) =\E_n[\{g(\mu_n(x) + C_n(x)Z_k) - f_n^*\}^+],
\end{equation}
where $C_n(x)$ is the lower Cholesky  factor of $K_n(x,x)$, and $Z_k$ is a $k$-dimensional standard normal random vector. (\shortciteN{astudillo2019bayesian} refers to \eqref{eq:ei_rep_trick} as the expected improvement for composite functions (EI-CF) to distinguish it from the classical expected improvement.) Equation \eqref{eq:ei_rep_trick} is then used to show that, under mild regularity conditions, $\EI_n$ is differentiable almost everywhere and its gradient, when it exists, is given by $\nabla_{x}\EI_n(x) = \E_n[\gamma_{n}(x;Z)]$, where
\begin{equation*}
    \gamma_n(x;Z_k) = \begin{cases}
\nabla_x g(\mu_{n}(x) + C_{n}(x)Z_k), \textnormal{ if } g(\mu_{n}(x) + C_{n}(x)Z_k) >  f_n^*,\\
0, \textnormal{ otherwise.}
\end{cases}
\end{equation*}
In particular, $\gamma_n$ provides an unbiased estimator of the gradient of $\EI_n$ which can be used within SGA with multiple restarts to maximize $\EI_n$. 
\savelength{
This approach is analogous to the one proposed by \shortciteN{wang2016parallel} for maximizing a batch version of EI that also lacks a closed-form analytical expression.
}

As a second approach, \shortciteN{balandat2020botorch} notes that  \eqref{eq:ei_rep_trick} also unlocks the adoption of a SAA scheme to approximately maximize $\EI_n$. This approach consists of fixing $M$ samples from a $k$-dimensional standard normal distribution, $Z_k^{(1)}, \ldots, Z_k^{(M)}$, and considering the Monte Carlo (MC) estimate of $\EI_n$ given by
\begin{equation*}
    \widehat{\EI}_n\left(x;Z_k^{(1:M)}\right) = \frac{1}{M} \sum_{m=1}^M \left\{g\left(\mu_n(x) + C_n(x)Z_k^{(m)}\right) - f_n^*\right\}^+.
\end{equation*}
Then, the problem
\begin{equation}
\label{eq:saa}
   \max_{x\in\X}\widehat{\EI}_n\left(x;Z_k^{(1:M)}\right) 
\end{equation}
is solved, and its solution is used as a proxy for the maximizer of $\EI_n$.

Importantly, since the samples $Z_k^{(1)}, \ldots, Z_k^{(M)}$ are fixed, \eqref{eq:saa} is a deterministic optimization problem. Moreover, under mild regularity conditions, $\widehat{\EI}_n\left(\cdot;Z_k^{(1:M)}\right)$ is differentiable, allowing the use of deterministic gradient-based optimization. \shortciteN{balandat2020botorch} shows empirically that this approach produces better results than SGA with multiple restarts at a lower computational cost. In addition, it shows that, under suitable regularity conditions, any solution of \eqref{eq:saa} converges in probability exponentially fast to a maximizer of $\EI_n$ as $M\rightarrow\infty$, thus suggesting that in practice it is safe to use low values of $M$.

While the discussion has focused on extending EI, it is possible to extend other acquisition functions following similar approaches. For example, \shortciteN{balandat2020botorch} derives an extension of the KG acquisition function for composite objective functions following an analogous SAA approach.

\subsection{Other Related Work}
The approach of \shortciteN{astudillo2019bayesian} was recently extended by \shortciteN{astudillo2021bayesian} to a more general class of composite objectives, evaluated via a series of functions, arranged in a directed acyclic network so that each function in the network takes as input the output of its parent nodes. Composite objective functions have also been considered outside the BO framework. For example \shortciteN{wild2017chapter} developed a trust-region method for derivative-free optimization of a composite objective function where  $g(y)=\|y - y^{\textnormal{obs}}\|_2^2$. In contrast with BO, this method is designed for local rather than global optimization. There is also a broad literature on gradient-based methods for optimizing composite objective functions \shortcite{burke1995gauss,shapiro2003class,drusvyatskiy2019efficiency}. In addition to derivatives, these methods often rely on convexity. Finally, composite (a.k.a. nested) functions have also been considered in GP-based sequential design of experiments with the goal of prediction rather than optimization \shortcite{marque2019efficient}.

\section{MULTI-FIDELITY BAYESIAN OPTIMIZATION}
\label{sec:multifidelity-BO}
\begin{figure}[h]
\floatbox[{\capbeside\thisfloatsetup{capbesideposition={right,top},capbesidewidth=8cm}}]{figure}[\FBwidth]
{\caption{
Multi-fidelity output for a steady-state queuing control problem, plotting objective value (total cost) versus the decision variable (service rate) at a collection of time horizons. Our goal is to choose the service rate to minimize the total cost at the longest time horizon pictured. Shorter time horizons offer approximations to the objective with less computational effort. Computational effort is approximately proportional to the time horizon.
\label{fig:mf_example}
}\label{fig:test}}
{\includegraphics[width=0.4\textwidth]{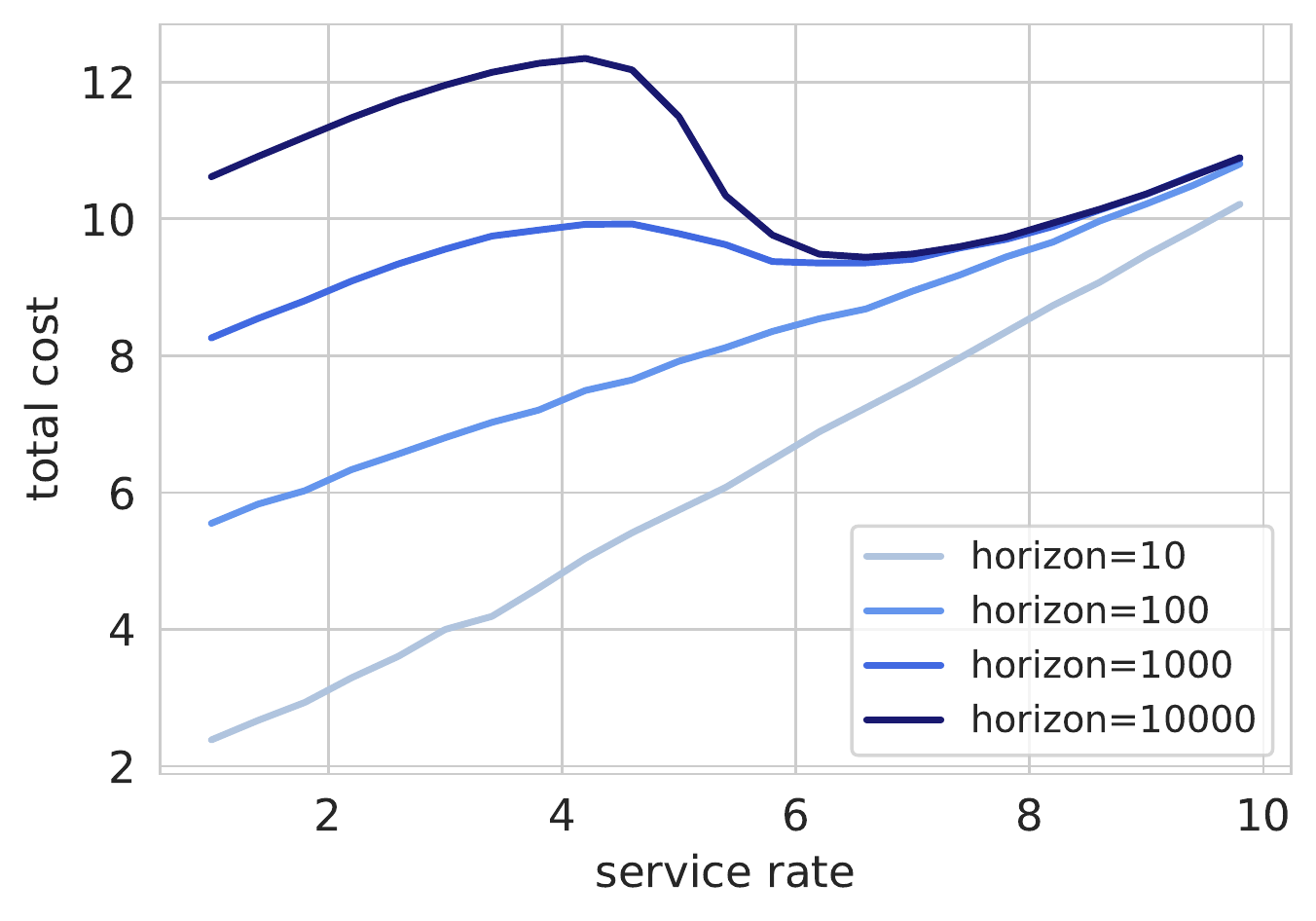}}
\end{figure}

The second grey-box setting we consider is multi-fidelity BO, where it is possible to evaluate cheaper approximations of the objective function by varying evaluation oracle parameters. This arises, for example, when optimizing steady-state peraformance as estimated by simulating over a long time horizon: we can simulate over a shorter time to quickly approximate the objective (see Figure \ref{fig:mf_example}). It also arises in hyperparameter tuning of DNNs trained via stochastic gradient descent (SGD): we can run SGD for a small number of iterations and obtain a proxy of the accuracy of the DNN that would result if it were trained until convergence using a potentially larger number of iterations. \savelength{In this case, in addition, we get to observe not only the accuracy of the model at the last iteration but also at every iteration of the SGD routine.} 
\savelength{Finally, it also arises in the calibration of expensive simulators that require numerically solving a system of complex PDEs: we can reduce the size of the grid in which the system of PDEs is solved to obtain a faster-to-evaluate approximation of the objective function.}

The multi-fidelity BO problem is formalized by assuming that the objective function is given by $f(x)=h(x,w^{\mathrm{tf}})$, where $h:\X\times\W\rightarrow\R$ is a black-box function,$\W$ is the space of fidelity-control parameters, and $w^{\mathrm{tf}}$ is the \textit{target fidelity}. We also assume that evaluating $h$ at $(x,w)$ has a cost $c(x,w)$. When the function $c:\X\times\W\rightarrow(0,\infty)$ is unknown (i.e., a black box), a common choice is to model $\log c$ using a GP.

\subsection{Predictive Model}
To effectively leverage the information provided by cheaper approximations of the objective function, it is key to use a predictive model  able to capture the correlation between these approximations and the objective function itself. As an illustrative example, below we discuss the model proposed by \shortciteN{poloczek2017multi}, which is as a special case of the \textit{semi-parametric latent factor model} \shortcite{alvarez2012kernels}.

Suppose that $\W$ is finite, say $\W=\{w_1,\ldots, w_k\}$. Making a slight abuse of notation, let $h(x)=\left(h(x,w_j):j=1,\ldots,k\right)$, and assume that $h_k$ is the objective to optimize (i.e., $w^\mathrm{tf} = w_k$). A sensible modeling choice is then to assume that $h_k$ and $h_j - h_k$ for $j=1,\ldots, k-1$ are drawn from independent GPs, which implicitly assumes that the bias of lower fidelity approximations and the objective function are all independent. This translates in a multi-output GP model over $h$ with prior mean and covariance functions of the form
\begin{equation*}
    \mu_0(x) = (\nu_0^{(1)}(x) + \mu_0^{(k)}(x), \ldots,  \nu_0^{(k-1)}(x) + \mu_0^{(k)}(x), \ \mu_0^{(k)}(x))
\end{equation*}
\begin{equation*}
    K_0(x,x') = \left(\mathbb{I}\{j=j' \textnormal{ and } j\neq k\}\Xi_0^{(j)}(x,x') + K_0^{(k)}(x,x')\right)_{j,j' = 1,\ldots, k},  
\end{equation*}
where $\nu_0^{(j)}$ and $\Xi_0^{(j)}$ are the prior mean and covariance functions of $h_j - h_k$, for $j=1,\ldots, k-1$, and $\mu_0^{(k)}$ and $K_0^{(k)}$ are the prior mean and covariance functions of $h_k$.

\subsection{Acquisition Functions}
Since the EI acquisition function is defined via a quantity that can only be measured when the evaluation of the objective function is performed fully (namely, the  improvement that would be obtained from such evaluation), it is difficult to extend it in a principled way to multi-fidelity evaluations.  Despite this difficulty, previous work (e.g., 
\shortciteNP{huang2006sequential}) has supported multi-fidelity extensions of EI
via heuristic arguments. 
However, other acquisition functions can be naturally extended to this setting. \shortciteN{wu2019practical}, for example, extends the KG acquisition function 
to measure incremental reward per unit cost,
\begin{equation*}
    \KG_n(x,w) = \E_n\left[\frac{\mu_{n+1}^* - \mu_n^*}{c(x_{n+1},w_{n+1})}\mid(x_{n+1}, w_{n+1})=(x,w)\right],
\end{equation*}
where $\mu_m^* = \max_{x\in\X}\E_m[f(x)]=    \max_{x\in\X}\mu_m(x, w^{\mathrm{tf}})$ for $m=n, n+1$.  

Other acquisition functions have been extended to the multi-fidelity evaluations setting.
\shortciteN{swersky2013multi}, for example, extends the ES acquisition function by dividing the expected information gain that would be obtained from evaluating a pair $(x_{n+1},w_{n+1})$ by its cost $c(x_{n+1},w_{n+1})$. An analogous extension of the max-value entropy search acquisition function was proposed by \shortciteN{takeno2020multi}. 

\subsection{Other Related Work}
Multi-fidelity optimization, including both Bayesian approaches and those that use non-Bayesian surrogate models, is the longest thread of literature of those grey-box problem classes considered in this tutorial.  Early work in this area focused on problems in engineering design, especially those where PDE mesh size could be controlled, and includes
\shortciteN{huang2006sequential} and \shortciteN{forrester2007multi}. 
This work also owes a great deal to non-optimization-focused work that developed surrogate models using multi-fidelity computer codes \shortcite{kennedy2000predicting}.
More recently, while interest in the use of multi-fidelity methods for engineering design (especially in aerospace) has been sustained (see, e.g., \shortciteNP{peherstorfer2018survey}), interest has also grown in its use in machine learning, especially for hyperparameter optimization \shortcite{wu2019practical,takeno2020multi}.
This remains an exciting area with great potential for value delivered through intelligent application of existing methods and the development of new methods and theory, especially aligned with the challenges of novel application domains.

\section{BAYESIAN OPTIMIZATION WITH OBJECTIVE CONSTITUENT EVALUATIONS}
\label{sec:constituent-BO}
Evaluation of just some of the objective function's constituents is possible in both BO of composite objective functions and multi-fidelity BO, and allows learning from these partial evaluations (and optionally later continuing paused evaluations). This strategy is already implicit in our discussion of multi-fidelity BO. For example, consider  
multi-fidelity BO approaches for hyperparameter tuning of DNNs, where lower fidelities are obtained by using fewer training iterations to optimize the weights of the DNN. An evaluation can be paused at a small number of iterations, the information obtained thus far can be incorporated into the predictive model, and then it can be continued later if desired. We thus focus here on describing objective constituent evaluations in the context of composite objective functions, where only some constituents of the multivariate output combined to produce the objective are evaluated at each iteration.

To motivate constituent evaluations for composite objective functions,
suppose we seek to find a configuration of bicycle docks within a city to maximize the average number of trips taken in a bike-sharing system \shortcite{freund2019analytics}. A simulator takes historical demand (in the form of request times and desired origin and destination) and simulates bike availability and the number of trips taken. The objective is $\sum_j h(x,w_j)$, where $x$ is a candidate configuration of bike docks, $j$ indexes days from which historical demands are taken, $w_j$ contains historical context about the day (e.g., the amount of rainfall), and $h(x,w_j)$ is the number of trips taken on that day. 

This problem can be tackled using BO of composite objective functions, where $h(x) = (h(x,w_j) : j)$ is the inner function and the outer function $g(h(x)) = \sum_j h_j(x)$ is the sum of this vector. One could then apply approaches from  the previous section. There is an opportunity, however, to improve efficiency further. One can easily evaluate just one term in this sum, $h(x,w_j)$. This would be much faster than evaluating the entire sum and might give information nearly as useful for optimization. 
An algorithm could selectively evaluate just one term at a time in this sum (just one constituent, $h_j(x) = h(x,w_j)$) to identify promising values of $x$ before spending the effort to evaluate the whole sum. This saves substantial time if, e.g., the sum is over 100 terms and each term takes one hour to compute.

By placing a GP prior over $h$ that models its dependence on both $x$ and $w$, we can perform inference over $g(h(x))$. We can then use an acquisition function to value an additional evaluation of $h$ at a single pair $(x,w)$, toward the goal of solving $\max_x g(h(x))$. This is the approach taken for outer functions $g$ that are sums or integrals in \shortciteN{williams2000sequential}, \shortciteN{xie2012optimization}, and \shortciteN{toscano2018bayesian}; risk measures in \shortciteN{cakmak2020risk}, and \shortciteN{nguyen2021value}; and arbitrary functionals of a control-dependent PDF modeled using a spatial logistic GP in \shortciteN{gautier2021goal}.



\savelength{It is worth comparing this approach, at least for linear functions $g$, with the following MC approach: choose $i$ uniformly at random and then return $h(x,w_j)$ as a noisy unbiased observation of the objective. Indeed, this approach also allows smaller investments to quickly learn whether a particular $x$ is promising doing additional sampling to obtain a high-accuracy estimate. There is, however, much that this MC approach cannot achieve. First, it does not generalize easily to nonlinear $g$. Second, if $h(x,w)$ is smooth in $w$, allowing $h(x,w')$ to be predicted well from $h(x,w)$ at a $w$ close to $w'$, then inference based on Bayesian quadrature can be substantially more accurate than a MC approach, especially if the $w$ are selected carefully.}



\subsection{Predictive Model}
As argued earlier,  constituent evaluations  (recall, one objective constituent is $h(\cdot,w)$ for a single $w$) can often provide information nearly as good as complete evaluations of the objective function at a much lower cost. This is particularly true when the constituents are highly correlated, and, therefore, using a statistical model capable of capturing such correlation is paramount.

When $h(x,w)$ is continuous in both $x$ and $w$, 
it is convenient to simply model $h$ with  
a GP prior that uses a standard distance-based covariance function $K:(\X\times\W)\times(\X\times\W)\rightarrow\R$. 
This is true even if the objective is a sum that only depends on $h(x,w)$ at finitely many values of $w$.


When $h(x,w)$ lacks smoothness in $w$,  
an effective choice is to use a multi-output GP model with \textit{intrinsic corregionalization}, whose covariance function, $K:\X\times\X \rightarrow \R^{k\times k}$, is of the form $K(x,x') = \Sigma K'(x,x')$, where $\Sigma\in\R^{k\times k}$ is a positive definite  matrix, and $K':\X\times\X\rightarrow\R$ is a covariance function. The matrix $\Sigma$ can be estimated with the other hyperparameters of the GP model. Other popular multi-output GP models include the semi-parametric latent factor model, discussed before, and the \textit{linear corregionalization model}. See \shortciteN{alvarez2012kernels} for details.

Regardless of the choice of the covariance function, a key property that has been leveraged to develop efficient acquisition functions when the objective function is a sum or integral of the individual constituents is that the implied posterior distribution on the objective distribution is again a GP. More concretely, if $h$ is modeled using a multi-output GP with posterior mean function $\mu_n:\X\rightarrow\R^k$ and covariance function $K_n:\X\times\X \rightarrow \R^{k\times k}$, then, using elementary properties of the multivariate normal distribution, it can be shown that, for any fixed $p\in\R^k$, the implied posterior distribution on $f = p^\top h$ is a GP with mean function $p^\top\mu_n$ and covariance function $p^\top K_n p$. A similar statement holds true for integrals \shortcite{ohagan1991bayes}.

\subsection{Acquisition Functions}
As in the case of multi-fidelity evaluations, it is unclear how to extend EI to handle constituent evaluations in a composite objective framework.
However, several  attempts have been made in the literature. For example, when $f$ is a sum or integral of the components of $h$, \shortciteN{williams2000sequential} proposes to select the next point to evaluate, $x_{n+1}$, by maximizing a variant of the classical EI acquisition function: it is computed with respect to the implied GP posterior distribution on $f$, taking $f^*_n$ to be the best $f(x)$ across all previously evaluated $x$. All previously evaluated $x$ are included, even those for which $f(x)$ is not well-estimated because $h(x,w)$ has been evaluated for only one $w$.
Having chosen $x_{n+1}$, the $w_{n+1}$ determining the constituent $h(x_{n+1},w_{n+1})$ to evaluate is chosen to minimize the posterior variance of $f(x_{n+1})$  after this evaluation is complete. As argued by \shortciteN{toscano2018bayesian}, however, this policy is unsatisfactory.
First, $x_{n+1}$ is chosen without considering $w_{n+1}$, while the best choice should be made jointly across $x$ and $w$: when observing $h(x, w)$ at a single $w$ produces a significant variance reduction, this should increase our willingness to evaluate at this $x$.
Second, in discrete problems, once $h(x,w)$ has been evaluated once at each $x$ for at least one $w$, this variant of EI becomes identically zero, producing no guidance and potentially leading to a lack of consistency.

While EI does not have a natural extension allowing for constituent evaluations, other acquisition functions do.  For example, again for the case where $f$ is a sum or integral of the components of $h$, \shortciteN{toscano2018bayesian} proposes an extension of the KG acquisition function, derived following the same decision-theoretic approach. More concretely, this acquisition function is defined as
\begin{equation*}
    \KG_n(x,w) = \E_n\left[\mu_{n+1}^* - \mu_n^*\mid(x_{n+1}, w_{n+1})=(x,w)\right],
\end{equation*}
where $\mu_m^* = \max_{x\in\X}\E_m[f(x)]=    \max_{x\in\X}\sum_{j=1}^k\mu_m(x, w_j)$,
for $m=n, n+1$. Importantly, in contrast with \shortciteN{williams2000sequential}, this acquisition function chooses the pair $(x,w)$, representing the input and constituent to be evaluated, jointly in a one-step optimal way. In numerical experiments, this acquisition function delivers significantly superior performance to the approach proposed by \shortciteN{williams2000sequential} and other approaches that select $x$ and $w$ separately.

The above acquisition function can be maximized by virtually unmodified versions of the approaches used to maximize KG for standard BO discussed earlier. This is due to the linear nature of the transformation mapping $h$ to $f$, causing the GP distribution on $h$ to imply a GP distribution on $f$. However, in many settings, the transformation that maps $h$ onto $f$ is non-linear (see \S\ref{sec:constituent-BO-other}). For such settings, approaches similar to those described in 
\S\ref{sec:composite-BO} can be employed to efficiently maximize MC-based acquisition functions.
\subsection{Other Related Work}
\label{sec:constituent-BO-other}
 As mentioned earlier, a related line of work studies problems analogous to those discussed in this section, where the transformation that maps  $h$ onto $f$ is non-linear. Such transformations often arise when seeking solutions $x$ that poses some form of \textit{risk aversion} to variations in $w$. Concrete examples include optimization of worst-case performance  \shortcite{marzat2013worst,bogunovic2018adversarially}; distributionally-robust optimization \shortcite{kirschner2020distributionally}; and optimization of risk measures \shortcite{cakmak2020risk,nguyen2021value}.
 



\section{CONCLUSIONS AND DIRECTIONS FOR FUTURE WORK}
\label{sec:conclusion}
Grey-box BO trades generality for performance gains that are, in some cases, quite dramatic. Since grey-box BO is a young research area and grey-box BO methods are customized to a problem class, or even to an individual problem, there are many questions that remain open. We believe that the following research questions are ones that are particularly interesting and likely to bear fruit over the coming years. 
\begin{itemize}
\item \textbf{Applications}: Applying grey-box BO in important and novel application domains can provide significant benefits while at the same time inspiring new methodological questions. Calibration of simulators and inverse reinforcement learning are exciting areas where composite objective functions can clearly provide benefits. In addition, atomistic simulation of chemical systems \cite{gillespie2007stochastic} is computationally intensive, important, and likely amenable to grey-box BO.
\item \textbf{Methods for new problem classes}: Working on novel applications is likely to identify other broad classes of grey-box structures and novel methods that productively leverage such structures.
\item \textbf{Many constituents}: Existing BO methods  for objective constituent evaluations become slow when there are many constituents. There is an opportunity to add value by developing more computationally efficient methods, e.g., by leveraging a known correlation structure between constituents that allows for faster predictive computations \shortcite{maddox2021bayesian}, or by intelligently selecting which constituents to model individually and which to aggregate.
\item \textbf{Non-myopic BO}: Non-myopic BO, which has shown promise in standard BO \shortcite{jiang2020efficient}, is likely to unlock even more value in grey-box BO. In grey-box BO, constituent evaluations do not provide a direct myopic benefit, and it is thus important to use knowledge-gradient or other methods that look further ahead than expected improvement to derive value. Looking further ahead will allow a method to understand when several pieces of information together can provide much more value than any one piece of information individually. 

\item \textbf{Theoretical understanding of grey-box BO methods}: 
There is much to be done deepening our theoretical understanding of grey-box BO methods by deriving regret bounds, convergence rates, and  understanding how problem structure determines the value derived from a grey-box approach. For example, 
when does leveraging composite objective structure perform better than standard BO?
\end{itemize}

{\footnotesize
\bibliographystyle{wsc}
\bibliography{bibl}
}
\end{document}